\documentclass[11pt,a4paper]{article}
\usepackage[hyperref]{acl2019}
\usepackage{times}
\usepackage{latexsym}

\usepackage{amsmath}
\usepackage{amssymb}
\usepackage{extarrows}

\usepackage{mathabx}
\usepackage{mathtools}
\usepackage{color}
\usepackage{float}

\newcommand{\red}[1]{\textcolor{red}{#1}}

\newcommand{\semantics}[1]{[\![\mbox{\em  #1\/}]\!]}

\newcommand{\rem}[1]{\relax}

\newcommand{\bluecheck}{}%
\DeclareRobustCommand{\bluecheck}{%
  \tikz\fill[scale=0.4, color=green]
  (0,.35) -- (.25,0) -- (1,.7) -- (.25,.15) -- cycle;%
}

\newcommand{\redcolor}[1]{\textcolor{red}{#1}}

\newcommand{\upred}{\redcolor{\uparrow}}

\newcommand{\downred}{\redcolor{\downarrow}}

\usepackage[normalem]{ulem}

\usepackage{tikz}
\usetikzlibrary{shapes,arrows,matrix}

\newcommand{\ra}{$\rightarrow~$}

\newcommand{\SB}{\mathscr{S}}

\newcommand{\KB}{\mathscr{K}}

\usepackage{linguex}
\usepackage{cgloss4e}

\usepackage{tikz}
\usepackage{tikz-qtree}
\usepackage{tikz-qtree-compat}
\usetikzlibrary{positioning}
\usepackage{proof}
\usepackage{subcaption}
\usepackage{todonotes}
\usepackage[mathscr]{euscript}
\usepackage{multirow}
\usepackage{stmaryrd}

\usepackage{url}

\aclfinalcopy

\title{MonaLog: a Lightweight System for Natural Language Inference \\Based on Monotonicity} 

\author{
Hai Hu$^\dagger$ \quad
  Qi Chen$^\dagger$ \quad
  Kyle Richardson$^\ddagger$ \quad \\
  \textbf{Atreyee Mukherjee}$^\dagger$ \quad
  \textbf{Lawrence S. Moss}$^\dagger$  \quad
  \textbf{Sandra K\"{u}bler}$^\dagger$ \\
 $^\dagger$Indiana University, Bloomington, IN, USA\\
$^\ddagger$Allen Institute for Artificial Intelligence, Seattle, WA, USA\\
{\tt \{huhai,qc5,atremukh,lmoss,skuebler\}@indiana.edu}\\
{\tt kyler@allenai.org}
}

\date{}

\begin{document}
\maketitle
\begin{abstract}

We present a new logic-based inference engine  for natural language inference (NLI) called MonaLog, which is based on natural logic and the monotonicity calculus. In contrast to existing logic-based approaches, our system is intentionally designed to be as lightweight as possible, and operates using a small set of well-known (surface-level) monotonicity facts about quantifiers, lexical items and token-level polarity information. Despite its simplicity, we find our approach to be competitive with other logic-based NLI models on the SICK benchmark. %
We also use MonaLog in combination with the current state-of-the-art model BERT in a variety of settings, including
for compositional data augmentation.  We show that MonaLog is capable of generating large amounts of high-quality training data for BERT, improving its accuracy on SICK. 

\end{abstract}

\section{Introduction}

There has been rapid progress on natural language inference (NLI) in the last several years, due in large part to recent advances in neural modeling \cite{conneau2017supervised} and the introduction of several new large-scale inference datasets \cite{SICK,snli,mnli,khot2018scitail}. Given the high performance of current state-of-the-art models, there has also been interest in understanding the limitations of these models (given their uninterpretability) \cite{naik2018stress,mccoy2019right}, as well as finding systematic biases in benchmark datasets \cite{gururangan2018annotation,poliak2018hypothesis}. %

In parallel to these efforts, there have also been recent logic-based approaches to NLI \cite{mineshima2015emnlp,ccg2lambda,martinez2017,AbzianidzeLangPro,YanakaMMB18}, which take inspiration from linguistics. In contrast to early attempts at using logic \cite{bos2005recognising}, these approaches have proven to be more robust.   However they tend to use many rules and their output can be hard to interpret.
It is sometimes unclear whether the attendant complexity is justified, especially given that such models are currently far outpaced by data-driven models and are generally hard to hybridize  with data-driven techniques. %

In this work, we introduce a new logical inference engine called MonaLog, which is based on natural logic and work on monotonicity stemming from \newcite{vanBenthemEssays86}. In contrast to the logical approaches cited above, our starting point is different in that we begin with the following two questions: 1) what is the \emph{simplest} logical system that one can come up with to solve empirical NLI problems (i.e., the system with minimal amounts of primitives and background knowledge)?; and 2) what is the lower-bound performance of such a model? Like other approaches to natural logic \cite{maccartney2008coling,Angeli}, our model works by reasoning over surface forms (as opposed to translating to symbolic representations) using a small inventory of monotonicity facts about quantifiers, lexical items and token-level polarity \cite{hu2018polarity}; \emph{proofs} in the calculus are hence fully interpretable and expressible in ordinary language. Unlike existing work on natural logic, however, our model avoids the need for having expensive alignment and search sub-procedures \cite{maccartney2008phrase,stern2011confidence}, and relies on a much smaller set of background knowledge and primitive relations than \newcite{MacCartneyManning}.

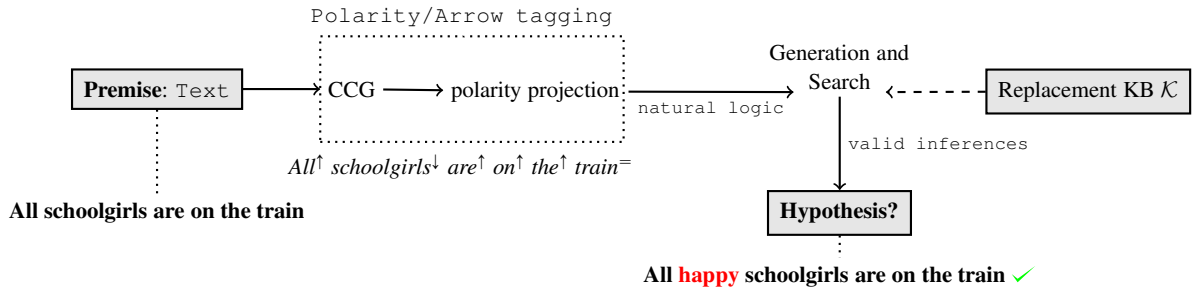
\begin{figure*}
\centering 
\definecolor{Gray}{gray}{0.9}
\begin{tikzpicture}[scale=0.7, every node/.style={scale=0.78}]
\tikzstyle{source} = [draw,thick,inner sep=.2cm,fill=Gray];
\tikzstyle{process} = [draw,thick,circle]; 
\tikzstyle{to} = [-,shorten >=1pt,semithick,font=\sffamily\footnotesize];
\tikzset{level distance=15pt,sibling distance=13pt,minimum size=7pt}'
\matrix (compiler) 
[matrix of nodes,
	row sep=1cm,
	column sep=.8cm,
	ampersand replacement=\&,
	nodes={align=left}
]
{
\node[source] (program) {\textbf{Premise}: \texttt{Text}}; \&[-4ex]  \node[](CCG) {\text{CCG}};  \& \node[](polarity) {\text{polarity projection}}; \&[-4ex]   \node[align=center,anchor=south](search) {\text{Generation and}  \\ \text{Search}};  \&[-9ex] \node[source](knowledge) {Replacement KB $\mathcal{K}$};  \\
\node[](girl) {\textbf{All schoolgirls are on the train}}; \& \& \& \node[source](hypo) {\textbf{Hypothesis?}}; \\[-.7cm] 
\& \& \& \node[](hypothesis) {\textbf{All \textcolor{red}{happy} schoolgirls are on the train }$\Large\bluecheck$}; \\[-1cm]
};
\draw[->,thick] (program) -- node[right] {\footnotesize } (CCG); 
\draw[->,thick] (CCG) -- node[right] {\footnotesize } (polarity);
\draw[->,thick] (polarity) -- node[below] {\footnotesize \texttt{natural logic}} ([xshift=-5ex,yshift=.5ex]search.south);
\draw[->,thick] (search) -- node[right] {\footnotesize \texttt{valid inferences}} (hypo);
\draw[dotted,thick] (CCG.west) ++(0,1) rectangle  ++(5.7,-2)   (polarity.north);
\draw[white] (polarity) -- ++(-1.4,1.1) node[above] {\textcolor{black}{\texttt{Polarity/Arrow tagging}}} (search);
\draw[white] (polarity) -- ++(-1.4,-1.8) node[above] {\textcolor{black}{\emph{All$^{\uparrow}$ schoolgirls$^{\downarrow}$ are${^{\uparrow}}$ on${^\uparrow}$ the$^{\uparrow}$ train${^=}$}}} (search);
\draw[->,dashed,thick] (knowledge) -- ([xshift=5ex,yshift=.5ex]search.south);
\draw[dotted,thick] (program) -- (girl); 
\draw[dotted,thick](hypo) -- (hypothesis);
\end{tikzpicture} 
\caption{An illustration of our general monotonicity reasoning pipeline using an example premise and hypothesis pair: \emph{All schoolgirls are on the train} and \emph{All happy schoolgirls are on the train}. \label{fig:pipeline} }
\end{figure*}

To show the effectiveness of our approach, we show results on the SICK dataset \cite{SICK},  a common benchmark for logic-based NLI, and find MonaLog to be competitive with more complicated logic-based approaches (many of which require full semantic parsing and more complex logical machinery). %
We also introduce a supplementary version of SICK that corrects several common annotation mistakes (e.g., asymmetrical inference annotations) based on previous work by \newcite{kalouli2017entail,kalouli2018}\footnote{Our correction can be found at: \urlstyle{rm}\url{https://github.com/huhailinguist/SICK_correction}}. Positive results on both these datasets show the ability of lightweight monotonicity models to handle many of the inferences found in current NLI datasets, hence putting a more reliable lower-bound on what results the simplest logical approach is capable of achieving on this benchmark.

Since our logic operates over surface forms, it is straightforward to hybridize our models. We investigate using MonaLog in combination with the %
language model BERT \cite{bert}, including for \emph{compositional data augmentation}, i.e, re-generating entailed versions of examples in our training sets. To our knowledge, our approach is the first attempt to use monotonicity for data augmentation, and we show that such augmentation can generate high-quality training data with which models like BERT can improve performance. 

\section{Our System: MonaLog\label{sec:sys:description}}

The goal of NLI is to determine, given a premise set $P$ and a hypothesis sentence $H$, whether $H$ follows from the meaning of $P$ \cite{Dagan}. In this paper, we  look at single-premise problems that involve making a standard 3-way classification decision (i.e., Entailment (H), Contradict (C) and Neutral (N)). Our general monotonicity reasoning system works according to the pipeline in Figure~\ref{fig:pipeline}. Given a premise text, we first do \texttt{Arrow Tagging} by assigning polarity annotations (i.e., the arrows $\uparrow,\downarrow$, which are the basic primitives of our logic) to tokens in text. These \emph{surface-level} annotations, in turn, are associated with a set of \texttt{natural logic} inference rules that provide instructions for how to generate entailments and contradictions  by span replacements over these arrows (which relies on a library of span replacement rules). For example, in the sentence \emph{All schoolgirls are on the train}, the token \emph{schoolgirls} is associated with a polarity annotation $\downarrow$, which indicates that in this sentential context, the span \emph{schoolgirls} can be replaced with a semantically more specific concept (e.g., \emph{happy schoolgirls}) in order to generate an entailment. A \texttt{generation and search} procedure is then applied to see if the hypothesis text can be generated from the premise using these inference rules. A \emph{proof} in this model is finally a particular sequence of edits (e.g., see Figure~\ref{fig:search_tree}) that derive the hypothesis text from the premise text rules and yield an entailment or contradiction. 

In the following sections, we provide the details of our particular implementation of these different components in MonaLog.

\subsection{Polarization (Arrow Tagging)}
Given an input premise $P$, MonaLog first polarizes each of its tokens and constituents, calling the system described by \citet{hu2018polarity}\footnote{\urlstyle{rm}\url{https://github.com/huhailinguist/ccg2mono}}, which performs polarization on a CCG parse tree. For example, a polarized $P$ could be \textit{every$^{\upred}$ linguist$^{\downred}$ swim$^{\upred}$}. Note that since we ignore morphology in the system, tokens are represented by lemmas. 

\subsection{Knowledge Base $\KB$ and Sentence Base $\SB$}

MonaLog utilizes two auxiliary sets. First, a knowledge base $\KB$ that stores the world knowledge needed for inference, e.g., \textit{semanticist} $\leq$ \textit{linguist} and \textit{swim} $\leq$ \textit{move}, which captures the facts that $\semantics{semanticist}$ denotes a subset of $\semantics{linguist}$,
and that 
$\semantics{swim}$
denotes a subset of $\semantics{move}$, respectively. Such world knowledge can be created manually for the problem at hand, or derived easily from existing resources such as WordNet \cite{wordnet}. 
Note that we do not blindly add \textit{all} relations from WordNet to our knowledge base, since this would hinge heavily on word sense disambiguation (we need to know whether the ``bank'' is a financial institution or a river bank to extract its relations correctly). In the current implementation, we avoid this by adding \textit{x} $\leq$ \textit{y} or \textit{x} $\perp$\footnote{$\perp$ means ``is contradictory to''.} \textit{y} relations only if both \textit{x} and \textit{y} are words in the premise-hypothesis pair.\footnote{There may be better and robust ways of incorporating WordNet relations to $\KB$; we leave this for future work. }
Additionally, some relations that involve quantifiers and prepositions need to be hard-coded, since WordNet does not include them: \textit{every} $=$ \textit{all} $=$ \textit{each} $\leq$ \textit{most} $\leq$ \textit{many} $\leq$ \textit{a few} $=$  \textit{several} $\leq$ \textit{some} $=$ \textit{a}; \textit{the} $\leq$ \textit{some} $=$  \textit{a}; \textit{on} $\perp$ \textit{off}; \textit{up} $\perp$ \textit{down}; etc. 

We also need to keep track of relations that can potentially be derived from the $P$-$H$ sentence pair. For instance, for all adjectives and nouns that appear in the sentence pair, it is easy to obtain: \textit{adj + n} $\leq$ \textit{n} (\textit{black cat} $\leq$ \textit{cat}). Similarly, we have \textit{n + PP/relative clause} $\leq$ n (\textit{friend in need} $\leq$ \textit{friend}, \textit{dog that bites} $\leq$ \textit{dog}), \textit{VP + advP/PP} $\leq$ \textit{VP} (\textit{dance happily/in the morning} $\leq$ \textit{dance}), and so on. We also have rules that extract pieces of knowledge from $P$ directly, e.g.: \textit{n$_1$} $\leq$ \textit{n$_2$} from sentences of the pattern \textit{every n$_1$ is a n$_2$}. One can also connect MonaLog to bigger knowledge graphs or ontologies such as DBpedia.

A sentence base $\SB$, on the other hand, stores the generated entailments and contradictions.

\subsection{Generation \label{sec:MonaLog:gensent}}

Once we have a polarized CCG tree, and some $\leq$ relations in $\KB$, generating entailments and contradictions is fairly straightforward. A concrete example is given in Figure~\ref{fig:search:tree}. Note that the generated $\leq$ instances are capable of producing mostly monotonicity inferences, but MonaLog can be extended to include other more complex inferences in \textit{natural logic}, hence the name Mo\textit{na}Log. This extension is addressed in more detail in \newcite{HuChenMoss}. 

\paragraph{Entailments/inferences}
The key operation for generating entailments is \texttt{replacement}, or substitution. It can be summarized as  follows: 1) For upward-entailing (UE) words/constituents, replace them with words/constituents that denote bigger sets. 2) For downward-entailing (DE) words/constituents, either replace them with those denoting smaller sets, or add modifiers (adjectives, adverbs and/or relative clauses) to create a smaller set. Thus for \textit{every$^{\upred}$ linguist$^{\downred}$ swim$^{\upred}$}, MonaLog can produce the following three entailments by replacing each word with the appropriate word from $\KB$: \textit{most$^{\upred}$ linguist$^{\downred}$ swim$^{\upred}$}, \textit{every$^{\upred}$ semanticist$^{\downred}$ swim$^{\upred}$} and \textit{every$^{\upred}$ linguist$^{\downred}$ move$^{\upred}$}.  These are results of one \texttt{replacement}.

Performing \texttt{replacement} for multiple rounds/depths can easily produce many more entailments. 

\paragraph{Contradictory sentences} To generate sentences contradictory to the input sentence, we do the following: 1) if the sentence starts with ``no (some)'', replace the first word with ``some (no)''. 2) If the object is quantified by ``a/some/the/every'', change the quantifier to ``no'', and vice versa. 
3) Negate the main verb or remove the negation. See examples in Figure~\ref{fig:search:tree}.

\paragraph{Neutral sentences} 
MonaLog returns Neutral if it cannot find the hypothesis $H$ in $\SB.entailments$ or $\SB.contradictions$. Thus, there is no need to generate neutral sentences. 

\subsection{Search}
\begin{figure*}[t]
	\centering
	\scalebox{0.80}{
		\begin{tikzpicture}
		\tikzset{align=center,level distance=45,sibling distance=30pt}
		\tikzset{edge/.style = {->,> = latex'}}
		\Tree
		[.{$P$: A$^{\upred}$ schoolgirl$^{\upred}$ with$^{\upred}$ a$^{\upred}$ black$^{\upred}$ bag$^{\upred}$ \\
		is$^{\upred}$ on$^{\upred}$ a$^{\upred}$ crowded$^{\upred}$ train$^{\upred}$
		} 
		[.\node(inf1){A$^{\upred}$     schoolgirl$^{\upred}$ is$^{\upred}$\\ on$^{\upred}$ a$^{\upred}$ crowded$^{\upred}$ train$^{\upred}$};  [.{...} ] [.{...} ] ] 
		[.\node(inf2){A$^{\upred}$ schoolgirl$^{\upred}$ with$^{\upred}$ a$^{\upred}$ bag$^{\upred}$ \\  is$^{\upred}$ on$^{\upred}$ a$^{\upred}$ crowded$^{\upred}$ train$^{\upred}$}; [.{...} ] [.{...} ] ]
		[.\node(inf3){ \red{A}$^{\upred}$ \red{girl}$^{\upred}$ \red{with}$^{\upred}$ \red{a}$^{\upred}$ \red{black}$^{\upred}$ \red{bag}$^{\upred}$ \\  \red{is}$^{\upred}$ \red{on}$^{\upred}$ \red{a}$^{\upred}$ \red{crowded}$^{\upred}$ \red{train}$^{\upred}$};
		[.{A girl \\ is on a crowded train} 
		[.{A girl is on a train} ]
		]
		]
		]
		\node [rectangle,draw,below left=2cm and 0.2cm of inf1] (contra1) {\footnotesize \textit{No} schoolgirl is \\ \footnotesize on a crowded train};
		\node [rectangle,draw,below left=2cm and 0cm of inf2] (contra2) {\footnotesize A schoolgirl with a bag \\ \footnotesize is \textit{not} on a crowded train};
		\node [rectangle,draw,below left=2cm and 0.2cm of inf3] (contra3) {\footnotesize ...};
		\draw[edge, shorten >=2pt, shorten <=2pt] (inf1) to (contra1);
		\draw[edge, shorten >=2pt, shorten <=2pt] (contra1) -- node[above, rotate=40] {\tiny contradiction} (inf1) ;
		\draw[edge, shorten >=2pt, shorten <=2pt] (inf2) to (contra2);
		\draw[edge, shorten >=2pt, shorten <=2pt] (contra2) -- node[above, rotate=40] {\tiny contradiction} (inf2) ;
		\draw[edge, shorten >=2pt, shorten <=2pt] (inf3) to (contra3);
		\draw[edge, shorten >=2pt, shorten <=2pt] (contra3) -- node[above, rotate=40] {\tiny contradiction} (inf3) ;
		\end{tikzpicture}
	}%
	\caption{Example search tree for SICK 340, where $P$ is \textit{A schoolgirl with a black bag is on a crowded train}, with the $H$: \red{\textit{A girl with a black bag is on a crowded train}}. Only one \texttt{replacement} is allowed at each step. Sentences at the nodes are generated entailments.
	\framebox{Sentences} in rectangles are the generated contradictions. In this case our system will return \texttt{entail}. The search will terminate after reaching the $H$ in this case, but for illustrative purposes, we show entailments of depth up to 3. To exclude the influence of morphology, all sentences are represented at the lemma level in MonaLog, which is not shown here.  \label{fig:search:tree} }
	\label{fig:search_tree}
\end{figure*}
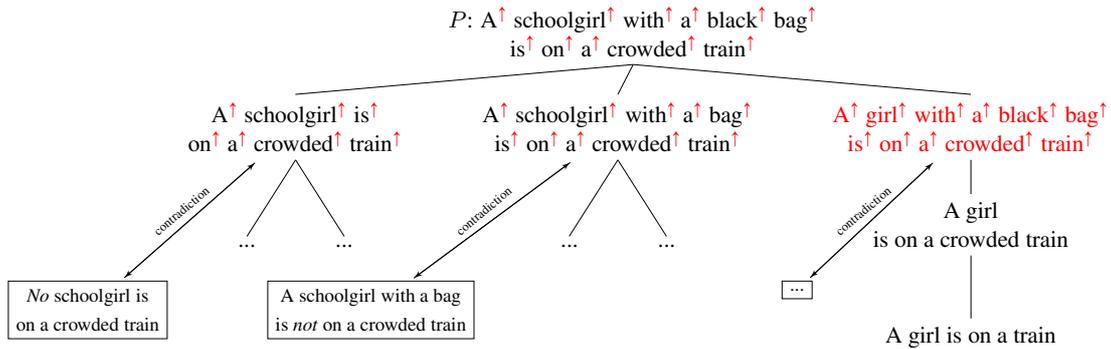

Now that we have a set of inferences and contradictions stored in $\SB$, we can simply see if the hypothesis is in either one of the sets by comparing the strings. If yes, then return Entailment or Contradiction; if not, return Neutral, as schematically shown in Figure~\ref{fig:search:tree}.
However, the exact-string-match method is too brittle. Therefore, we apply a heuristic. If the only difference between sentences $S_1$ and $S_2$ is in the set \{``a'', ``be'', ``ing''\}, then $S_1$ and $S_2$ are considered semantically equivalent. 

The search is implemented using depth first search, with a default depth of 2, i.e.~at most 2 replacements for each input sentence. 
At each node, MonaLog ``expands'' the sentence (i.e., an entailment of its parent)
by obtaining its entailments and contradictions, and checks whether $H$ is in either set. If so, the search is terminated; otherwise the systems keeps searching until all the possible entailments and contradictions up to depth 2 have been visited. 

\section{MonaLog and SICK\label{sec:monalog:and:sick}} 
We perform two experiments to test MonaLog. We first use MonaLog to solve the problems in a commonly used 
natural language inference dataset, SICK \cite{SICK}, 
comparing our results with previous systems. 
Second, we test the quality of the data generated by MonaLog. 
To do this, we generate more training data (sentence pairs) from the SICK training data using our system, and performe fine-tuning on BERT \cite{bert}, a
language model based on the transformer architecture \cite{vaswani2017attention}, 
with the expanded dataset. In all experiments, we use the Base, Uncased model of BERT\footnote{\urlstyle{rm}\url{https://github.com/google-research/bert}}.

\subsection{The SICK Dataset}

The SICK \cite{SICK} dataset includes around 10,000 English sentence pairs that are annotated to have either ``Entailment'', ``Neutral'' or ``Contradictory'' relations. 
We choose SICK as our testing ground for several reasons. 
First, we want to  test on a large-scale dataset, 
since we have shown that a similar model \cite{HuChenMoss} 
reaches good results on parts of the smaller FraCaS dataset \cite{fracas}. 
Second, we want to make our results comparable to those of previous logic-based models such as the ones described in \cite{bjerva2014,Abzianidze,martinez2017,YanakaMMB18}, which were also tested on SICK. 
We use the data split provided in the dataset:
4,439 training problems, 4,906 test problems and 495 trial problems, see Table~\ref{tab:sick} for examples. %

\begin{table*}[t]
		\begin{tabular}{l|p{5.9cm}|p{5.9cm}|rr}
		 & & & orig. &corr.\\
			id & premise & hypothesis &  label &  label  \\ 
			\hline
			219 & There is no girl in white dancing & A girl in white is dancing & C & C \\
			294 & Two girls are lying on the ground	& Two girls are sitting on the ground & N & C \\
			743 & A couple who have just got married are walking down the isle&	The bride and the groom are leaving after the wedding & E & N \\
			1645 & A girl is on a jumping car&One girl is jumping on the car&  E& N \\
			1981 & A truck is quickly going down a hill &	A truck is quickly going up a hill & N & C \\
			8399 & A man is playing guitar next to a drummer & A guitar is being played by a man next to a drummer & E & n.a. \\ \hline
		\end{tabular}
	\caption{Examples from SICK \cite{SICK} and corrected SICK \cite{kalouli2017entail,kalouli2018} w/ syntactic variations. n.a.:example not checked by Kalouli and her colleagues. C: contradiction; E: entailment; N: neutral.\label{tab:sick}}
\end{table*}

\subsection{Hand-corrected SICK}

There are numerous issues with the original SICK dataset, as illustrated by \citet{kalouli2017entail,kalouli2018}.

They first manually checked 1,513 pairs tagged as ``A entails B but B is neutral to A'' (\textit{AeBBnA}) in the original SICK, correcting 178 pairs that they considered to be wrong \cite{kalouli2017entail}. Later, \citet{kalouli2018} extracted pairs from SICK whose premise and hypothesis differ in only one word, and created a simple rule-based system that used WordNet information to solve the problem. Their WordNet-based  method was able to solve 1,651 problems, whose original labels in SICK were then manually checked and corrected against their system's output. They concluded that 336 problems are wrongly labeled in the original SICK.
Combining the above two corrected subsets of SICK, minus the overlap, results in their corrected SICK dataset\footnote{\urlstyle{rm}\url{https://github.com/kkalouli/SICK-processing}}, which has 3,016 problems (3/10 of the full SICK), with 409 labels different from the original SICK (see breakdown in Table~\ref{tab:corr:sick}). 16 of the corrections are in the trial set, 197 of them in the training set and 196 in the test set. This suggests that more than one out of ten problems in SICK are potentially problematic. For this reason, two authors of the current paper checked the 409 changes. We found that only 246 problems are labeled the same by our team and by \citet{kalouli2018}.
 For cases where there is disagreement, we adjudicated the differences after a discussion. 

\begin{table}[t]
    \centering
    \begin{tabular}{c|c|c|c|c}
        total & N \ra E & E \ra C & N \ra C & E \ra N \\\hline
        409 & 14 & 7 & 190 & 198 \\ \hline
    \end{tabular}
    \caption{Changes from SICK to corrected SICK \cite{kalouli2017entail,kalouli2018}.  \label{tab:corr:sick}}
\end{table}

We are aware that the partially checked SICK (by two teams) is far from ideal. We therefore present results for two versions of SICK for experiment 1 (section \ref{sec:exp1}): the original SICK and the version  corrected by our team. For the data augmentation experiment in section \ref{sec:exp2}, we only performed fine-tuning on the corrected SICK. As shown in a recent SICK annotation experiment by \newcite{kalouli2019explaining}, annotation is a complicated issue influenced by linguistic and non-linguistic factors. We leave checking the full SICK dataset to future work.

\section{Experiment 1: Using MonaLog Directly\label{sec:exp1}}

\subsection{Setup and Preprocessing}

The goal of experiment 1 is to test how accurately MonaLog solves problems in a large-scale dataset. We first used the system to solve the 495 problems in the trial set and then manually identified the cases in which the system failed. Then we determined which syntactic transformations are needed for MonaLog. 
After improving the results on the trial data by introducing a preprocessing step to handle limited syntactic variation (see below), we applied MonaLog on the test set. This means that the rule base of the system was optimized on the trial data, and we can test its generalization capability on the test data. 

The main obstacle for MonaLog is the syntactic variations in the dataset, illustrated in some examples in Table~\ref{tab:sick}. 
There exist multiple ways of dealing with these variations: One approach is to `normalize' unknown syntactic structures to a known structure. For example, we can transform passive sentences into active ones and convert existential sentences into the base form (see ex.~8399 and 219 in Table~\ref{tab:sick}).
Another approach is to use some more abstract syntactic/semantic representation so that the linear word order can largely be ignored, e.g., represent a sentence by its dependency parse, or use Abstract Meaning Representation. Here, we explore the first option and leave the second approach to future work. %
We believe that dealing with a wide range of syntactic variations requires tools designed specifically for that purpose. The goal of MonaLog is to generate entailments and contradictions based on a polarized sentence instead.

Below, we list the most important syntactic transformations we perform in preprocessing\footnote{For the complete list of transformations see: \urlstyle{rm}\url{https://github.com/huhailinguist/SICK_correction}}.

\begin{enumerate}
  \setlength{\itemsep}{1pt}
  \setlength{\parskip}{0pt}
  \setlength{\parsep}{0pt}
    \item Convert all passive sentences to active using \textit{pass2act}\footnote{\urlstyle{rm}\url{https://github.com/DanManN/pass2act}}. If the passive does not contain a \textit{by} phrase, we add \textit{by a person}.
	\item  Convert existential clauses into their base form (see ex.~219 in Table~\ref{tab:sick}).
	\item Other transformations: \textit{someone/anyone/no one} \ra \textit{some/any/no person}; \textit{there is no man doing sth.} \ra \textit{no man is doing sth.}; etc. 
\end{enumerate}

\subsection{Results}

 \begin{table}[t]
\scalebox{0.85}{
\begin{tabular}{c|rrr}
	system & P & R & acc. \\ \hline
    \multicolumn{4}{c}{On \textbf{uncorrected} SICK} \\ \hline
	majority baseline & -- & -- & 56.36 \\
	hypothesis-only baseline & \multirow{2}{*}{--} & \multirow{2}{*}{--} & \multirow{2}{*}{56.87} \\
	\cite{poliak2018hypothesis} &&& \\\hline
	\multicolumn{4}{c}{MonaLog (this work)} \\\hline
	MonaLog + all transformations & 83.75  & 70.66 & 77.19 \\ 
	Hybrid: MonaLog + BERT  &  83.09 &  85.46 & 85.38  \\
    \hline
	\multicolumn{4}{c}{ML/DL-based systems} \\\hline
	BERT (base, uncased) & 86.81 & 85.37 & 86.74 \\
	\cite{GRU} & -- & -- & \textbf{87.1} \\ 
	\cite{beltagy2016} & -- & -- & 85.1 \\
	\hline
	\multicolumn{4}{c}{Logic-based systems} \\\hline
	\cite{bjerva2014} & 93.6 & 60.6 & 81.6 \\
	\cite{Abzianidze} & 97.95 & 58.11& 81.35 \\
	\cite{martinez2017} & 97.04 & 63.64 & 83.13 \\
	\cite{YanakaMMB18} & 84.2 & 77.3 & 84.3 \\
	\hline
	\multicolumn{4}{c}{ } \\
	\multicolumn{4}{c}{On \textbf{corrected} SICK} \\ \hline
	MonaLog + existential trans. & 89.43 & 71.53 & 79.11 \\
	MonaLog + pass2act & 89.42 & 72.18 & 80.25 \\
	MonaLog + all transformations & 89.91  & 74.23 & 81.66 \\ 
\hline
	Hybrid: MonaLog + BERT  &85.65&87.33& \textbf{85.95} \\
	BERT (base, uncased) & 84.62 & 84.27 & 85.00 \\ \hline
\end{tabular}
}
\caption{Performance on the SICK test set, original SICK above and corrected SICK below. 
P / R for MonaLog averaged across three labels. Results involving BERT are averaged across six runs; same for later experiments.  \label{tab:res}}
\end{table}

The results of our system on uncorrected and corrected SICK are presented in Table~\ref{tab:res}, along with comparisons with other systems.

Our accuracy on the uncorrected SICK (77.19\%) is much higher than the majority baseline (56.36\%) or the hypothesis-only baseline (56.87\%) reported by \citet{poliak2018hypothesis}, and only several points lower than current logic-based systems. %
Since our system is based on \textit{natural logic}, there is no need for translation into logical forms, which makes the reasoning steps transparent and much easier to interpret.
I.e., with entailments and contradictions, we can generate a natural language trace of the system, see Fig.~\ref{fig:search:tree}.

Our results on the corrected SICK are even higher (see lower part of Table~\ref{tab:res}), demonstrating the effect of data quality on the final results. Note that with some simple syntactic transformations we can gain 1-2 points in accuracy.

Table~\ref{tab:precrec} shows MonaLog's performance on the individual relations. The system is clearly very good at identifying entailments and contradictions, as demonstrated by the high precision values, especially on the corrected SICK set (98.50 precision for E and 95.02 precision for C). The lower recall values are due to MonaLog's current inability to handle syntactic variation. 

Based on these results,  we tested a hybrid model of MonaLog and BERT (see Table~\ref{tab:res}) where we exploit MonaLog's strength: Since MonaLog has a very high precision on Entailment and Contradiction, we can always trust MonaLog if it predicts E or C; when it returns N, we then fall back to BERT. 
This hybrid model improves the accuracy of BERT by 1\% absolute to 85.95\% on the corrected SICK.
On the uncorrected SICK dataset, the hybrid  system performs worse than BERT.

Since MonaLog is optimized for the corrected SICK, it may mislabel many E and C judgments in the \textit{uncorrected} dataset. %

The stand-alone BERT system performs better on the uncorrected data (86.74\%) than the corrected set (85.00\%). The corrected set may be too inconsistent since only a part has been checked.

Overall, these hybird results show that it is possible to combine our high-precision system with deep learning architectures. However, more work is necessary to optimize this combined system.

\begin{table*}[t]
\centering
\begin{tabular}{l|rr|rr|rr}
 & \multicolumn{2}{|l}{E} & \multicolumn{2}{|l}{C} & \multicolumn{2}{|l}{N}  \\
 & P & R & P & R & P & R \\ \hline
uncorr.\ SICK & 97.75 & 46.74 & 80.06 & 70.24 & 73.43 & 94.99  \\ 
corr.\ SICK & 98.50 & 50.46 & 95.02 & 73.60 & 76.22 & 98.63  \\
\hline
\end{tabular}
\caption{Results of MonaLog per relation. C: contradiction; E: entailment; N: neutral.}
\label{tab:precrec}
\end{table*}

\begin{table*}[t]
\begin{tabular}{r|p{4.8cm}|p{4.6cm}|l|l|l}
	id & premise & hypothesis & SICK & corr.~SICK & Mona \\ \hline 
	359 & There is no dog chasing another or holding a stick in its mouth & Two dogs are running and carrying an object in their mouths & N & n.a. & C\\ %
	1402 & A man is crying&	A man is screaming & N & n.a. & E \\ %
	1760 & A flute is being played by a girl & There is no woman playing a flute & N &  n.a.  & C \\ %
	2897 & The man is lifting weights & The man is lowering barbells & N & n.a.  & E \\ %
	2922 & A herd of caribous is not crossing a road &  A herd of deer is crossing a street & N & n.a.  & C \\ %
	3403 & A man is folding a tortilla & A man is unfolding a tortilla & N & n.a.  & C \\ %
	4333 & A woman is picking a can & A woman is taking a can & E & N & E \\ %
	5138 & A man is doing a card trick & A man is doing a magic trick & N & n.a.  & E  \\ %
	5793 & A man is cutting a fish & A woman is slicing a fish & N & n.a. & C \\ \hline
\end{tabular}
\caption{Examples of incorrect answers by MonaLog; %
n.a.\ = the problem has not been checked in corr.\ SICK. \label{tab:mistakes} }
\end{table*}

\subsection{Error Analysis}

Upon closer inspection, some of MonaLog's errors consist of difficult cases, as shown in Table~\ref{tab:mistakes}. For example, in ex.\ 359,
if our knowledge base $\KB$ contains the background fact
$\mbox{\em chasing} \leq \mbox{\em running}$, then MonaLog's judgment of C would be correct.
In ex.\ 1402, if \textit{crying} means \textit{screaming}, then the label should be E; however, if \textit{crying} here means \textit{shedding tears}, then the label should probably be N. 
Here we also see potentially problematic labels (ex.\ 1760, 3403) in the original SICK dataset. 

Another point of interest is that 19 of MonaLog's mistakes are related to the antonym pair \textit{man} vs.\ \textit{woman} (e.g., ex.~5793 in Table~\ref{tab:mistakes}). This points to inconsistency of the SICK dataset: Whereas there are at least 19 cases tagged as Neutral (e.g., ex.~5793), there are at least 17 such pairs that are annotated as Contradictions in the test set (e.g., ex.~3521), P: \textit{A man is dancing}, H: \textit{A woman is dancing} (ex.~9214), P: \textit{A shirtless man is jumping over a log}, H: \textit{A shirtless woman is jumping over a log}. If \textit{man} and \textit{woman} refer to the same entity, then clearly that entity cannot be \textit{man} and \textit{woman} at the same time, which makes the sentence pair a contradiction. If, however, they do not refer to the same entity, then they should be Neutral. %

\section{Experiment 2: Data Generation Using MonaLog\label{sec:exp2}}

Our second experiment focuses on using MonaLog to 
generate additional training data for machine learning models such as BERT.
To our knowledge, this is the first time that a rule-based NLI system has been successfully
used to generate training data for a deep learning application. %

\subsection{Setup}

As described above,  MonaLog generates entailments and contradictions when solving problems. These can be used as additional training data for a machine learning model.

I.e., we pair the newly generated sentences

with their input sentence, creating new pairs for training. For example, we take all the sentences in the \textit{nodes} in Figure~\ref{fig:search:tree} as inferences and all the sentences in \textit{rectangles} as contradictions, and then form sentence pairs with the input sentence.  The additional data can be used directly, almost without human intervention.

Thus for experiment 2, the goal is to examine the quality of these generated sentence pairs. For this, we re-train a BERT model on these pairs. 
If BERT trained on 
the manually annotated SICK training data
is improved by adding data generated by MonaLog,
then we can conclude that the generated data is of high quality, even comparable to human annotated data, which is what we found.

More specifically, we compare the performance of BERT models trained on a) SICK training data alone, and b) SICK training data plus the entailing and contradictory pairs generated by MonaLog.

All experiments are carried out using our corrected version of the SICK data set.

\begin{table*}[t]
    \centering
    \begin{tabular}{l|p{6cm}|p{6cm}|l}
        label & premise & hypothesis &comm. \\ \hline
        E & A woman be not cooking something & A person be not cooking something & correct \\ %
        E & A man be talk to a woman who be seat beside he and be drive a car & A man be talk & correct \\ %
        E & A south African plane be not fly in a blue sky & A south African plane be not fly in a very blue sky in a blue sky & unnat. \\ %
        C & No panda be climb	& Some panda be climb & correct \\ %
        C & A man on stage be sing into a microphone	& A man be not sing into a microphone & correct \\ %
        C & No man rapidly be chop some mushroom with a knife &	Some man rapidly be chop some mushroom with a knife with a knife & unnat. \\\hline 
        E & 
        Few$^{\upred}$ people$^{\downred}$ be$^{\downred}$ eat$^{\downred}$ at$^{\downred}$ red$^{\downred}$ table$^{\downred}$ in$^{\downred}$ a$^{\downred}$ restaurant$^{\downred}$ without$^{\downred}$ light$^{\upred}$
        & 
        Few$^{\upred}$ large$^{\downred}$ people$^{\downred}$ be$^{\downred}$ eat$^{\downred}$ at$^{\downred}$ red$^{\downred}$ table$^{\downred}$ in$^{\downred}$ a$^{\downred}$ Asian$^{\downred}$ restaurant$^{\downred}$ without$^{\downred}$ light$^{\upred}$
        & correct \\\hline
    \end{tabular}
    \caption{Sentence pairs generated by MonaLog, lemmatized. \label{tab:gen:sents} }
\end{table*}

However, note that MonaLog is designed to only generate entailments and contradictions. Thus, we only have access to newly generated examples for those two cases, we do not acquire any additional neutral cases. Consequently, adding these examples to the training data will introduce a skewing that does not reflect the class distribution in the test set. Since this will bias the machine learner against neutral cases, we use the following strategy to counteract that tendency: We relabel all cases where BERT is not confident enough
for either E or C into N. We set this threshold to 0.95 but leave further optimization of the threshold to future work.

\subsection{Data Filtering and Quality Control}

MonaLog is prone to over-generation. For example, it may wrongly add the same adjective before a noun (phrase) twice to create a more specific noun, e.g., \textit{young young man }$\leq$ \textit{young man} $\leq$ \textit{man}. Since it is possible that such examples influence the machine learning model negatively, we look into filtering such examples to improve the quality of the additional training data.

We manually inspected 100 sentence pairs generated by MonaLog to check the quality and naturalness of the new sentences (see Table~\ref{tab:gen:sents} for examples). 
All of the generated sentences are correct in the sense that the relation between the premise and the hypothesis is correctly labeled as entailment or contradiction (see Table~\ref{tab:manu}).

While we did not find any sentence pairs with wrong labels,  some generated sentences are unnatural, as shown in Table \ref{tab:gen:sents}. 
Both unnatural examples contain two successive copies of the same PP.

\begin{table}[t]
\centering
    \begin{tabular}{l|r|rrrr}
         label & total & correct & wrong & unnatural  \\\hline
         E & 56 & 49 & 0 & 7\\
         C & 44 & 41 &0 & 3 \\ \hline
    \end{tabular}
\caption{Quality of 100 manually inspected sentences.\label{tab:manu}}
\end{table}

Note that our data generation hinges on correct polarities on the words and constituents. For instance, in the last example of Table~\ref{tab:gen:sents}, the polarization system needs to know that \textit{few} is downward entailing on both of its arguments, and \textit{without} flips the arrow of its argument, in order to produce the correct polarities, on which the \texttt{replacement} of MonaLog depends.

In order to filter unnatural sentences, such as the examples in Table~\ref{tab:gen:sents}, we use a rule-based filter and remove sentences that contain bigrams of repeated words\footnote{We also investigated using a bigram based language model to filter out non-natural sentences. However, this affected the results negatively.}. We experiment with using one quarter  or one half randomly selected sentences in addition to a setting where we use the complete set of generated sentences.

\subsection{Results}

\begin{table*}[t]
\begin{center}
		\begin{tabular}{l|rrr|r}
			training data & \# E & \# N & \# C & acc. \\\hline
			SICK.train: baseline & 1.2k & 2.5k & 0.7k & 85.00 \\\hline
			1/4 gen. + SICK.train & 8k & 2.5k & 4k & 85.30 \\
			1/2 gen. + SICK.train & 15k & 2.5k & 7k & 85.81 \\
			all gen. + SICK.train & 30k & 2.5k & 14k & 86.51 \\ 
			E, C prob.~threshold = 0.95 & 30k & 2.5k & 14k & 86.71 \\\hline
	Hybrid baseline &  1.2k & 2.5k & 0.7k & 85.95 \\			
	Hybrid + all gen. & 30k & 2.5k & 14k & 87.16 \\
		Hybrid + all gen. +	threshold & 30k & 2.5k & 14k & \textbf{87.49} \\\hline
		\end{tabular}
	\caption{Results of BERT trained on MonaLog-generated entailments and contradictions plus SICK.train (using the corrected SICK set).
	\label{tab:res:gen:w:neutral}
	}
\end{center}
\end{table*}

Table~\ref{tab:res:gen:w:neutral} shows the amount of additional sentence pairs per category along with the results of using the automatically generated sentences as additional training data. %

It is obvious that adding the additional training data results in gains in accuracy even though the training data becomes increasingly skewed towards E and C. When we add all additional sentence pairs, accuracy increases by more than 1.5 percent points. This demonstrates both the robustness of BERT in the current experiment and the usefulness of the generated data.  The more data we add, the better the system performs. 

We also see that raising the threshold to relabel uncertain cases as neutral gives a small boost, from 86.51\% to 86.71\%. This translates into 10 cases where the relabeling corrected the answer.%

Finally, we also investigated whether the hybrid system, i.e.,  MonaLog followed by the re-trained BERT, can also profit from the additional training data. Intuitively, we would expect smaller gains since MonaLog already handles a fair amount of the entailments and contradictions, i.e., those cases where BERT profits from more examples. However the experiments show that the hybrid system reaches an even higher accuracy of 87.16\%, more than 2 percent points above the baseline, equivalent to roughly 100 more problems correctly solved. 
Setting the high threshold for BERT to return E or C further improves accuracy to 87.49\%. This brings us into the range of the state-of-the-art results, even though a direct comparison is not possible because of the differences between the corrected and uncorrected dataset.

\section{Conclusions and Future Work}

We have presented
a working natural-logic-based system, MonaLog, which attains high accuracy
on the SICK dataset
and can be used to generated natural logic proofs.
Considering how simple and straightforward our method is, we believe it can serve as a strong baseline or basis for other (much) more complicated systems, either logic-based or ML/DL-based. In addition, we have shown that MonaLog can
generate high-quality training data, which improves the accuracy of a deep learning model when trained on the expanded dataset. 
As a minor point, we manually checked the corrected SICK dataset by \citet{kalouli2017entail,kalouli2018}. 

There are several directions for future work. The first direction concerns the question how to handle syntactic variation from natural language input. That is, the computational process(es) for inference will usually be specified in terms of strict
syntactic conditions, and  naturally occurring sentences will typically not
conform to those conditions. 
Among the strategies  which allow their systems to better cope with premises and hypotheses with various syntactic structures are sophisticated versions of alignment used by e.g.~\newcite{MacCartney,YanakaMMB18}. We will need to extend MonaLog to be able to handle such variation. In the future, we plan to use dependency relations as representations of natural language input and train a classifier that can determine which relations are crucial for inference.

Second, as mentioned earlier, we are in need of a fully (rather than partially) checked SICK dataset to examine the impact of data quality on the results since the partially checked dataset may be inherently inconsistent between the checked and non-checked parts.

Finally, with regard to the machine learning experiments, we plan to investigate other methods of addressing the imbalance in the training set created by additional entailments and contradictions. 
We will look into options for artificially creating neutral examples, e.g.~by finding reverse entailments\footnote{In the set relations by \newcite{MacCartney}, if $A \sqsubset B$, then $A$ entails $B$, but $B$ is neutral to $A$.}, as illustrated by \newcite{richardson2019probing}. 

\section*{Acknowledgements}
We thank the anonymous reviewers for their helpful comments. Hai Hu is supported by China Scholarship Council.

\bibliography{MonaLog_arxiv_clean}
\bibliographystyle{acl_natbib}

\end{document}